\documentclass[conference,a4paper]{IEEEtran}
\usepackage{cite}
\usepackage{amsmath,amssymb,amsfonts}
\usepackage{graphicx}
\usepackage{textcomp}
\usepackage{xcolor}
\def\BibTeX{{\rm B\kern-.05em{\sc i\kern-.025em b}\kern-.08em
    T\kern-.1667em\lower.7ex\hbox{E}\kern-.125emX}}

\newcommand\skipthis[1]{}

\date{\today}

\begin{document}

\title{Pre or Post-Softmax Scores in Gradient-based Attribution Methods, What is Best?}

\skipthis{
\author{\IEEEauthorblockN{1\textsuperscript{st} Given Name Surname}
\IEEEauthorblockA{\textit{dept. name of organization (of Aff.)} \\
\textit{name of organization (of Aff.)}\\
City, Country \\
email address or ORCID}
\and
\IEEEauthorblockN{2\textsuperscript{nd} Given Name Surname}
\IEEEauthorblockA{\textit{dept. name of organization (of Aff.)} \\
\textit{name of organization (of Aff.)}\\
City, Country \\
email address or ORCID}
}
} 

\author{\IEEEauthorblockN{Miguel Lerma}
\IEEEauthorblockA{\textit{Department of Mathematics} \\
\textit{Northwestern University}\\
Evanston, Illinois, USA \\
mlerma@math.northwestern.edu \\ ORCID: 0000-0003-4703-6417}
\and
\IEEEauthorblockN{Mirtha Lucas}
\IEEEauthorblockA{\textit{College of Computing and Digital Media} \\
\textit{DePaul University}\\
Chicago, Illinos, USA \\
mlucas3@depaul.edu \\ ORCID: 0000-0002-4252-7746}
}

\IEEEoverridecommandlockouts
\IEEEpubid{\makebox[\columnwidth]{979-8-3503-3337-4/23/\$31.00 ©2023 IEEE \hfill}
\hspace{\columnsep}\makebox[\columnwidth]{ }}

\maketitle

\IEEEpubidadjcol

\begin{abstract}
  Gradient based attribution methods for neural networks working as
  classifiers use gradients of network scores.  Here we discuss the
  practical differences between using gradients of pre-softmax scores
  versus post-softmax scores, and their respective advantages and
  disadvantages.
\end{abstract}


\section{Introduction}

In the last few years the area of eXplainable Artificial Intelligent
(XAI) has gained increasing attention.  For deep neural networks,
explanation methods often take the form of attribution algorithms 
determining the impact of each input feature on a given output.

In particular, gradient-based attribution methods work by computing
the gradient
$\nabla_{\mathbf{x}}S = (\partial S/\partial x_1,\dots,\partial
S/\partial x_N)$ of an output or ``score'' $S$ of the network respect
to a set of inputs or unit activations $\mathbf{x} = (x_1,\dots,x_N)$.
The assumption is that each derivative $\partial S/\partial x_i$
provides a measure of the impact of $x_i$ on the score~$S$.

Classifier networks such as the VGG family \cite{simonyan2015}
typically have a final layer with as many outputs as classes
(\figurename\,\ref{f:dnn}).  The class imputed to an input sample is given by
the score with the largest value.  In those kinds of networks it is
common to place at the end a \emph{softmax} activation function
defined as
\begin{equation}\label{e:softmaxdef}
  y_c = \frac{e^{z_c}}{\sum_{i=1}^n e^{z_i}}
  \,,
\end{equation}
where $z_1,\dots,z_n$ are the (pre-softmax) outputs of the last layer.
Then, the final (post-softmax) outputs $y_1,\dots,y_n$ of the network
form a vector of positive scores that add to 1, so they can be
interpreted as a distribution of probabilities.  The network is
trained with a loss function that depends on its post-softmax outputs
and ground-truth target values.

Gradient-based attribution methods use different choices
concerning whether to use gradients of pre-softmax $z_i$, or
gradients of post-softmax $y_i$ scores.  The following are a few examples:

\begin{itemize}

\item As described in \cite{selvaraju2019}, Grad-CAM uses the
  gradients of pre-softmax scores, although we have found
  implementations in which post-softmax scores are used instead
  \skipthis{(see e.g. \cite{rosebrock2020,meudec2021,uddin2021}),}
  \skipthis{(see e.g. \cite{rosebrock2020,uddin2021}),}
  (see e.g. \cite{uddin2021}),
  which makes sense if we consider that those are the scores compared to the
  target scores during training. A problem with using post-softmax
  scores in Grad-CAM is that their gradients tend to vanish when the
  outputs are close to saturation.

\item Integrated Gradients (IG) \cite{sundararajan2017ig} cannot use
  pre-softmax scores without losing the property of being model-agnostic,
  so IG is bound to use post-softmax scores only.

\item As described in \cite{lucas2022rsi}, RSI Grad-CAM uses post-softmax
  scores.  Unlike Grad-CAM, this method does not suffer from the
  vanishing gradients problem because it uses gradients from a
  sequence of interpolating inputs intended to capture the total
  change of the gradient from a baseline to the given network input.
  While some of the gradients in intermediate steps of the
  interpolation may be zero, the total change of gradients from
  baseline is less likely to also be zero.

\skipthis{
\item Grad-CAM++ \cite{chattopadhyay2018art} and Grad-CAM plus \cite{lerma2022}.
The authors of Grad-CAM++ require the pre-softmax scores to be fed to
a smooth function, which could be a softmax, although the implementation is
much simpler when using just an element-wise exponential function $y^c = e^{z^c}$ 
(this is the implementation we have seen used in practice).
However, in this case \cite{lerma2022} shows that Grad-CAM++ becomes
practically equivalent to an even simpler small variation 
of the original Grad-CAM that uses only positive gradients,
called Grad-CAM$^+$ (Grad-CAM~plus) in the paper.
}

\item Grad-CAM++ \cite{chattopadhyay2018art} and Grad-CAM plus \cite{lerma2022}.
The implementation of Grad-CAM++ has pre-softmax scores fed to
an element-wise exponential function $y^c = e^{z^c}$.
In \cite{lerma2022} it is shown that Grad-CAM++ is
practically equivalent to a small variation 
of the original Grad-CAM that uses only positive gradients,
called Grad-CAM$^+$ (Grad-CAM~plus) by the authors.
\end{itemize}

Further examples of gradient-based attribution methods can be found in \cite{ancona2022}.

\skipthis{
\begin{figure*}[htb]
\centering
\ \includegraphics[height=1.8in]{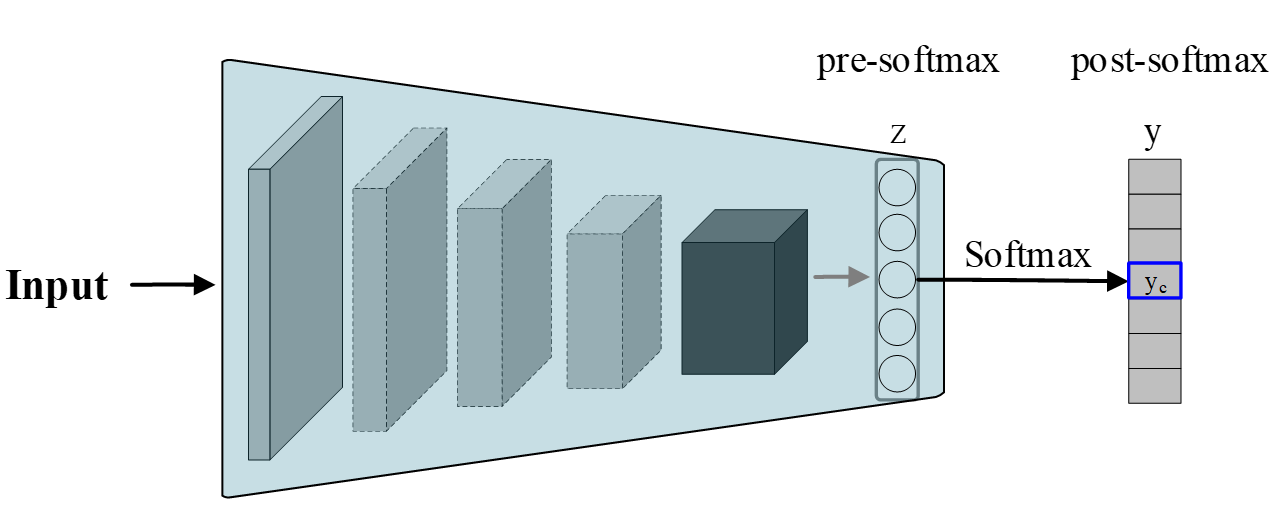}
\caption{Structure of a typical classifier network. After a number of
  convolutional blocks this kind of network ends with a fully
  connected network producing a (pre-softmax) output $z$,
  followed by a softmax activation function with (post-softmax) output
  $y$.}\label{f:dnn}
\end{figure*}
}

\begin{figure}[htb]
\centering
\ \includegraphics[width=3.45in]{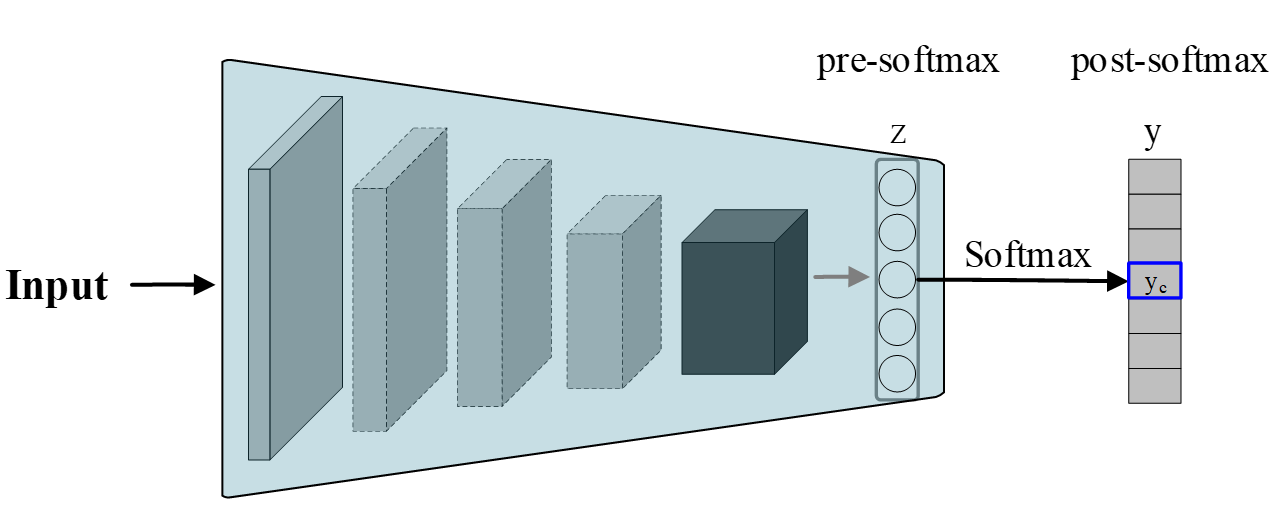}
\caption{Structure of a typical classifier network. After a number of
  convolutional blocks this kind of network ends with a fully
  connected network producing a (pre-softmax) output $z$,
  followed by a softmax activation function with (post-softmax) output
  $y$.}\label{f:dnn}
\end{figure}

\section{Pre-softmax vs post-softmax outputs }

Understanding the main differences between using gradients of pre and
post-softmax outputs requires first to look at how gradients
backpropagate through a softmax. Then, we will discuss how each choice
is related to the functioning of a network that has been trained by
minimizing a given loss function.

\subsection{Gradient of a function in terms of pre and post-softmax
  scores.}\label{s:fun}
Differentiating the softmax function defined in (\ref{e:softmaxdef})
with respect to $z_i$ we get
\begin{equation}
\begin{aligned}
  \frac{\partial y_c}{\partial z_i} &=
  \frac{\frac{\partial}{\partial z_i} e^{z_c}}{\sum_{j=1}^n e^{z_j}}
  - \frac{e^{z_c} \frac{\partial}{\partial z_i} \sum_{j=1}^n e^{z_j}}{(\sum_{j=1}^n e^{z_j})^2} \\
  &= \frac{e^{z_c} \delta_{ic}}{\sum_{j=1}^n e^{z_j}}
  - \frac{e^{z_c} e^{z_i}}{(\sum_{i=1}^n e^{z_i})^2}
  = y_c (\delta_{ic} - y_i)
\end{aligned}
\end{equation}
where $\delta_{ij}$ is Kronecker delta, defined
\begin{equation}
  \delta_{ij} = \begin{cases}
    1 & \text{ if } i=j \,, \\
    0 & \text{ if } i \neq j \,.
    \end{cases}
\end{equation}
Hence, the derivative of the softmax function is
\begin{equation}\label{e:diffsoftmax}
    \frac{\partial y_c}{\partial z_i} = y_c (\delta_{ic} - y_i)
    \,,
\end{equation}

If $f(y_1,\dots,y_n)$ is a (differentiable) function of the
post-softmax outputs of the network, and $\mathbf{x} = (x_i,\dots,x_N)$
is a set of network inputs or activations of hidden units, we can find
the gradient $\nabla_\mathbf{x} f$ by using the chain rule:
\begin{equation}\label{e:L1}
  \nabla_\mathbf{x} f =
  \sum_{c} \frac{\partial f}{\partial y_c} \nabla_\mathbf{x} y_c
  \,.
\end{equation}
Using the chain rule again we get
\begin{equation}\label{e:L2}
    \nabla_\mathbf{x} f =
  \sum_{c,i} \frac{\partial f}{\partial y_c} \frac{\partial y_c}{\partial z_i} \nabla_\mathbf{x} z_c
  = \sum_{c,i} \frac{\partial f}{\partial y_c} y_c (\delta_{ic} - y_i) \nabla_\mathbf{x} z_i
  \,.
\end{equation}
Equation (\ref{e:L1}) shows the gradient of $f$ in terms of gradients
of post-softmax scores, and (\ref{e:L2}) does it in terms of gradients
of pre-softmax scores.

\subsection{Impact of pre and post-softmax score gradients in
  detecting a class}\label{s:out}

Assume we want to determine the impact of a given unit activation or
input variable $x$ in detecting a class $c$.  Letting $f = y_c$, equation
(\ref{e:L1}) becomes a trivial identity
$\frac{\partial y_c}{\partial x} = \frac{\partial y_c}{\partial x}$,
and (\ref{e:L2}) becomes
\begin{equation}\label{e:L3}
  \frac{\partial y_c}{\partial x} = y_c \sum_{i} (\delta_{ic} - y_i) \frac{\partial z_i}{\partial x}
  \,.
\end{equation}
We immediately see that there can be situations in which the pre and
post-softmax score gradients may lead to radically different
conclusions about the impact of a given activation in the process of
class detection.  First, recalling that
$y_c = {e^{z_c}}/{\sum_{i=1}^n e^{z_i}}$, we see that replacing $z_i$
with $z'_i = z_i + t$, where $t$ is independent of $i$, we also
have $y_c = {e^{z'_c}}/{\sum_{i=1}^n e^{z'_i}}$ because:
\begin{equation}
\begin{aligned}
  \frac{e^{z'_c}}{\sum_{i=1}^n e^{z'_i}} &= \frac{e^{z_c+t}}{\sum_{i=1}^n e^{z_i+t}} \\
  &= \frac{e^t \, e^{z_c}}{e^t \sum_{i=1}^n e^{z_i}}
  = \frac{e^{z_c}}{\sum_{i=1}^n e^{z_i}} = y_c
  \,.
\end{aligned}
\end{equation}
So, the change $z_i \mapsto z_i+t$ for every $i$ does not change the
network post-softmax outputs $y_c$.  It does not change the
post-softmax gradients either, as can be seen by replacing $z_i$ with
$z'_i = z_i+t$ in equation (\ref{e:L3}):
\begin{equation}
  \begin{aligned}
    \frac{\partial y_c}{\partial x} &= y_c \sum_{i} (\delta_{ic} - y_i) \frac{\partial z'_i}{\partial x}
    = y_c \sum_{i} (\delta_{ic} - y_i) \frac{\partial (z_i + t)}{\partial x} \\
  &= y_c \sum_{i} (\delta_{ic} - y_i) \frac{\partial z_i}{\partial x} +
  y_c \frac{\partial t}{\partial x} \underbrace{\sum_{i} (\delta_{ic} - y_i)}_{0} \\
  &= y_c \sum_{i} (\delta_{ic} - y_i) \frac{\partial z_i}{\partial x} 
   \,,
   \end{aligned}
\end{equation}
where we have used $\sum_{i} \delta_{ic} = 1$, and $\sum_{i} y_i = 1$,
hence $\sum_{i} (\delta_{ic} - y_i) = 1-1=0$.  However,
$\partial z'_i/\partial x - \partial z_i/\partial x = \partial
t/\partial x$, hence, even though the post-softmax outputs $y_c$ and
the post-softmax gradients $\partial y_c/\partial x$ remain the same,
the pre-softmax score gradients $\partial z'_i/\partial x$ and
$\partial z_i/\partial x$ may be very different.  In particular it is
possible that two different trainings of the network may produce two
different models for which outputs and post-softmax gradients are the
same, while the pre-softmax gradients are very different.  In such
situation saliency maps using post-softmax gradients would be the
same, but the ones obtained using pre-softmax gradients would be very
different even though the two models are locally functionally equivalent
(in a neighborhood of a given input).

For another situation in which pre and post-softmax gradients may
yield radically different results is as follows.  Assume that
${\partial z_i}/{\partial x}$ is the same for all $i$, i.e.,
${\partial z_i}/{\partial x}=K$, for $i=1,\dots,n$. Then
\begin{equation}
  \frac{\partial y_c}{\partial x} = y_c \sum_{i} (\delta_{ic} - y_i) K =
   y_c (1 - 1) K = 0
  \,,
\end{equation}
where we have again used $\sum_{i} \delta_{ic} = 1$, and
$\sum_{i} y_i = 1$.  Hence, using gradients of post-softmax scores we
would conclude that $x$ has no impact in the detection of class $c$
for the particular network input used.  However, for the pre-softmax
gradients we have ${\partial z_i}/{\partial x}=K$, which can be
anything, large or small.  If we have two different activations $x_1$
and $x_2$ such that ${\partial z_i}/{\partial x_1}=K_1$ and
${\partial z_i}/{\partial x_2}=K_2$ for every~$i$, then
${\partial y_c}/{\partial x_1} = {\partial y_c}/{\partial x_2} = 0$,
implying that $x_1$ and $x_2$ have the same null impact in the final
output corresponding to class~$c$.  However,
${\partial z_i}/{\partial x_1}=K_1$ and
${\partial z_i}/{\partial x_2}=K_2$ may have very different values and
lead to a very different gradient-based saliency map compared to the
one we would obtain using gradients of post-softmax scores.  What is
worse, taking into account that different trainings may produce
(locally) equivalent functional models with very different pre-softmax
gradients, it is perfectly possible that those different training may
lead to situations in which $K_1 \gg K_2$, with the conclusion that
$x_1$ has a much larger contribution than $x_2$ in the detection of
class~$c$, and also situations in which $K_1 \ll K_2$, and the
opposite conclusion would hold.

In the next section we will examine the impact of pre and post-softmax
score gradients on the loss function.

\subsection{Impact of pre and post-softmax score gradients on the loss
  function}

The most common loss function for training classifier networks with a
final softmax is \emph{cross entropy}:
\begin{equation}\label{L1}
  \mathcal{L} = - \sum_{c=1}^n t_c \log y_c
  \,,
\end{equation}
where $y_c$ and $t_c$ are the softmax output and target output for
class $c$ respectively.  The cross entropy function is rooted on
information theory, and reaches its minimum precisely when $y_c=t_c$
for all classes $c$, i.e.,
$\min \mathcal{L} = - \sum_{c=1}^n t_c \log t_c$.  The difference
between $\mathcal{L}$ and $\min \mathcal{L}$ can be interpreted as the
information gained when the predicted class probability distribution
$y_c$ is replaced with the actual distribution $t_c$.  Given an
attribution method it is natural to determine in what extent the
attributions assigned have an impact on the information gain of the
predicted class distribution, which can be measured by the gradient of
the cross entropy loss function $\nabla_{\mathbf{x}} \mathcal{L}$.
This gradient can be computed by following the steps outlined in
section~\ref{s:fun} with $f = \mathcal{L}$.

In classification tasks with the target given as a 1-hot vector we
have $t_c = \delta_{c\bar{c}}$, where $\bar{c}$ is the ground-truth
class.  In this case $t_{\bar{c}} = 1$, and the loss function can
be written $\mathcal{L} = - \log y_{\bar{c}}$. The more general
expression (\ref{L1}) plays a role in cases in which the ground truth
cannot be expressed as a 1-hot vector, e.g. when an input may belong
to more than one class.

The partial derivative of the loss function w.r.t. $y_c$ is
\begin{equation}
  \frac{\partial \mathcal{L}}{\partial y_c} = - \frac{t_c}{y_c}
  \,,
\end{equation}
and its derivative w.r.t. $z_c$ is
\begin{equation}
\begin{aligned}
  \frac{\partial \mathcal{L}}{\partial z_c} &=
  \sum_{c'} \frac{\partial \mathcal{L}}{\partial y_{c'}} \frac{\partial y_{c'}}{\partial z_c} \\
  &= - \sum_{c'}  \frac{t_{c'}}{y_{c'}} y_{c'} (\delta_{c{c'}} - y_c) \\
  &= - t_c + y_c \sum_{c'} t_{c'} = y_c - t_c
  \,.
\end{aligned}
\end{equation}
In the last step we have used $\sum_c t_c = 1$.

Let $x$ represent the activation of a hidden unit or network input.
We can compare the roles of the pre-softmax and post-softmax partial
derivatives $\partial z_i/\partial x$ and $\partial y_i/\partial x$ in
the computation of the gradient of the loss function as follows.
Applying the chain rule we have:
\begin{equation}
  \frac{\partial \mathcal{L}}{\partial x} = \sum_{c}
  \frac{\partial \mathcal{L}}{\partial z_c} \frac{\partial z_c}{\partial x}
  = - \sum_c (t_c - y_c) \frac{\partial z_c}{\partial x}
  \,,
\end{equation}
and
\begin{equation}
\frac{\partial \mathcal{L}}{\partial x} = \sum_{c}
\frac{\partial \mathcal{L}}{\partial y_c} \frac{\partial y_c}{\partial x}
= - \sum_c \frac{t_c}{y_c} \frac{\partial y_c}{\partial x}
\,.
\end{equation}
In order to decouple prediction errors from explanation errors we can
focus only on data samples for which the model yields the right
predictions, so assume that $c$ is the right class associated to the
given input, and $(t_1,\dots,t_n)$ is a 1-hot vector,
$t_{c'} = \delta_{cc'}$.  Then
\begin{equation}\label{e:pre}
  \frac{\partial \mathcal{L}}{\partial x} 
  = - \sum_{c'} (\delta_{cc'} - y_{c'}) \frac{\partial z_{c'}}{\partial x}
  \,,
\end{equation}
and
\begin{equation}\label{e:post}
  \frac{\partial \mathcal{L}}{\partial x}
= - \frac{1}{y_{c}} \frac{\partial y_{c}}{\partial x}
\,.
\end{equation}

So, we see that the gradient of the pre-softmax score $z_c$ does not
capture the whole impact of $x$ on the loss function, for that we
would need the gradients of all the pre-softmax scores $z_{c'}$,
$c'=1,\dots,n$.  On the other hand, the gradient of the post-softmax
score $y_c$ alone captures the impact of $x$ on the loss function,
while the gradients of the other post-softmax scores $y_{c'}$
($c'\neq c$) have no effect.  From $\mathcal{L} = - t_{c} \log y_{c}$,
the relation between gradient of loss function and gradient of
post-softmax score can also be written:
\begin{equation}
  \frac{\partial y_{c}}{\partial x} = - \frac{\partial \exp(\mathcal{L})}{\partial x}
  \,,
\end{equation}
stressing the interpretation of the gradient of each post-softmax
score as a function of the loss function.  No similar relation exists
between each $\frac{\partial z_{c'}}{\partial x}$ and
$\frac{\partial \mathcal{L}}{\partial x}$ because equation
(\ref{e:pre}) is not invertible in general.

\skipthis{
\begin{figure}[htb]
\centering
\ \includegraphics[height=2in]{gradcam}
\ \includegraphics[height=2in]{rsi-gradcam}
\caption{Various saliency maps generated for Grad-CAM and RSI Grad-CAM
  respectively at layer \texttt{block5\_pool} of a VGG19 network
  pretrained on ImageNet, using logits (pre-softmax), post-softmax, and
  log-softmax scores respectively. The differences between using plain
  post-softmax and log-softmax scores are barely visible. Grad-CAM
  does a slightly better job at locating each flower when using
  pre-softmax scores.  RSI~Grad-CAM successfully locates each flower
  independently in all cases.}\label{f:example}
\end{figure}
}

\begin{figure}[htb]
\centering
\ \includegraphics[width=3.4in]{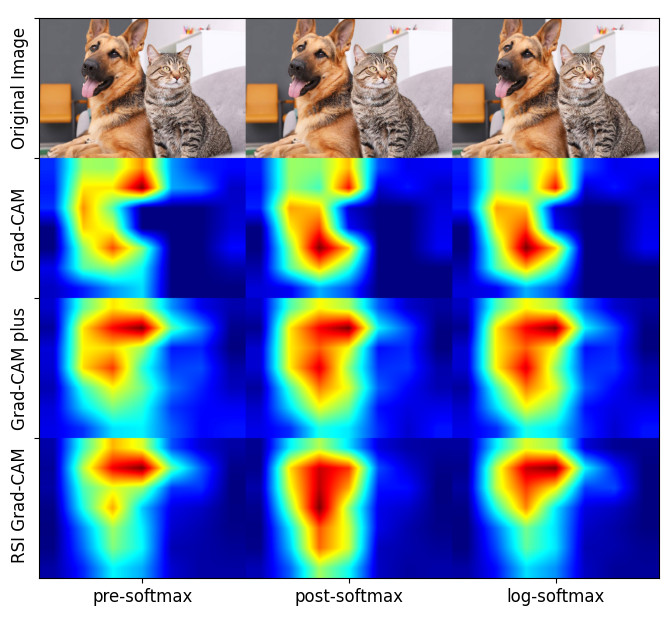}
\caption{Various saliency maps generated for Grad-CAM, Grad-CAM~plus,
and RSI Grad-CAM
  respectively at layer \texttt{block5\_pool} of a VGG19 network
  pretrained on ImageNet, using pre-softmax, post-softmax, and
  log-softmax scores respectively. In this example the differences between using
  post-softmax and log-softmax scores are barely visible for Grad-CAM and Grad-CAM~plus,
  however they make a difference for RSI~Grad-CAM.}\label{f:example}
\end{figure}

\subsection{Log-softmax scores.}

Equation (\ref{e:post}) can be rewritten
\begin{equation}\label{e:post2}
\frac{\partial \mathcal{L}}{\partial x}
= - \frac{\partial \log{y_{c}}}{\partial x}
\,.
\end{equation}
This brings up the question of whether gradients of the \emph{log-softmax} scores $\log{y_{c}}$ could also be used in a
gradient-based attribution methods since they have a more direct
relation to information gain.  To test this idea we performed some
preliminary tests with gradient-based attribution methods
(particularly Grad-CAM and RSI~Grad-CAM) using gradients of 
log-softmax scores ($\partial \log{y_{c}}/\partial x$), but with few exceptions
they did not yield results that were noticeable different from the ones
obtained using plain post-softmax scores (see \figurename\,\ref{f:example}
for an illustrative example).  However, it may still be worth it to
test them in new gradient-based attribution methods in the future.

\section{Discussion}

We can look at the outputs of a classifier network in two different ways:

\begin{enumerate}
\item \label{op1} The intended outputs are the pre-softmax scores, with
  the post-softmax scores being just a convenient, user-friendly way to
  represent the network output as a probability distribution.
\item \label{op2} The intended outputs are the post-softmax scores, with
  the pre-softmax scores being just an intermediate necessary step to
  compute the final (post-softmax) scores.
\end{enumerate}

The outcome of a classification is the same in both cases if we set
the class selected by the model as the one corresponding to the
maximum score.  However, to be consistent, approach~(\ref{op1}) would
require to use a loss function written in terms of the pre-softmax
scores and targets representing the desired outputs of said
pre-softmax scores.  The closest implementation of this idea we are
aware of is given by the \emph{softmax cross entropy with logits} loss
function, which in practice is implemented by performing internally a
softmax on the pre-softmax scores (logits) and then applying the usual
cross entropy with post-softmax targets, so it is in fact approach
(\ref{op2}) in disguise.

Hence, the gradients that best capture the impact of a sample input
on a given class output are the post-softmax gradients, in the
following sense:

\begin{itemize}
\item The impact of the sample can be interpreted as a measure of how
  the increase in the intensity of each input or activation produces
  an information gain increase.

  \newpage 
  
\item That impact can be channeled through a single class output
  rather than a combination of all class outputs.
\item The paradoxical situation discussed in section~\ref{s:out} in
  which equally well trained networks may lead to radically different
  saliency maps is less likely to happen.
\end{itemize}

\section{Conclusions}

We have discussed the advantages and disadvantages of using
pre-softmax versus post-softmax scores with gradient-based attribution
methods for classifier networks.  We have shown arguments in favor of
generally using post-softmax scores, with some possible exceptions.
In general, methods that are prone to the vanishing gradients problem
(such as Grad-CAM and some of its derivatives) may do better by using
pre-softmax scores, while models that are not affected by that problem
(like the ones based on path integrals of gradients such as Integrated
Gradients and RSI~Grad-CAM) may yield more robust results by using
post-softmax scores.  Finally, we suggest the use of log-softmax
scores as a third alternative that may deserve attention.

\bibliographystyle{IEEEtranS}
\bibliography{mybibfile}

\end{document}